\documentclass{article}
\usepackage{spconf,amsmath,graphicx,color,amssymb, float}

\title{Transfer Learning for Retinal Vascular Disease Detection: A Pilot Study with Diabetic Retinopathy and Retinopathy of Prematurity}
\name{Guan Wang$^{\star}$ \qquad Yusuke Kikuchi\thanks{${}^\S$ Corresponding author. Email: yusuke\_kikuchi@berkeley.edu}$^{\S\dagger}$ \qquad Jinglin Yi$^{\ddag}$ \qquad Qiong Zou$^{\ddag}$ \qquad Rui Zhou$^{\ddag}$ \qquad Xin Guo$^{\dagger}$}
\address{$^{\star}$ Tsinghua-Berkeley Shenzhen Institute, China\\
      $^{\dagger}$ University of California, Berkeley, USA\\
      $^{\ddag}$ Affiliated Eye Hospital of Nanchang University, China} 
\begin{document}
%
\maketitle
\begin{abstract}
Retinal vascular diseases affect the well-being of human body and sometimes provide vital signs of otherwise undetected bodily damage. Recently, deep learning techniques have been successfully applied for detection of diabetic retinopathy (DR). The main obstacle of applying deep learning techniques to detect most other retinal vascular diseases is the limited amount of data available. 

In this paper, we propose a transfer learning technique that aims to utilize the feature similarities for detecting retinal vascular diseases. We choose the well-studied DR detection as a source task and identify the  early  detection  of  retinopathy  of prematurity (ROP) as the target task.  
Our experimental results demonstrate that our  DR-pretrained approach dominates in all metrics the conventional ImageNet-pretrained transfer learning approach, currently adopted in medical image analysis. Moreover, our approach is more  robust with respect to  the stochasticity in the training process and with respect to reduced training samples. 

This study suggests the potential of our proposed transfer learning approach for a broad range of retinal vascular diseases or pathologies, where data is limited.

\end{abstract}
\section{Introduction}
\label{sec:intro}
\subsection{Problem}
The health of eyes is beyond being simply an integral part of the well-being  of human body. 
In particular, retinal vascular disease, referring to a condition that affects the blood vessels of the
retina, is well recognized to provide early
signals of bodily damage and is sometimes the only symptom of a person with a serious cardiovascular condition \cite{alanazi_osuagwu_almubrad_ahmed_ogbuehi_2015}. This is because  the arrangement of blood vessels at the back of the eye, known as the retinal vasculature, is closely connected to the health of heart \cite{Cheung2013}.
Diabetic retinopathy (DR) is another example of retinal vascular disease, with the lesions of retinal blood vessels caused by diabetes complications
\cite{atlas2015international, Schaneman2010}.

 Recently, deep learning techniques have been applied to detecting retinal vascular diseases. The most notable success is detection of  DR
\cite{10.1001/jama.2016.17216}.
 One of the key reasons behind this success is the vast amount of available data sets, often in the order of several hundred thousands, for both the negative and positive images of DR. Indeed, developing high-performance deep learning algorithms for medical image analysis generally requires collections of large data sets with tens of thousands of abnormal (positive) cases.
 
 Unfortunately, 
 data sets for most retinal vascular diseases are far more limited (often less than several thousands), and  generally imbalanced between negative and positive images. This is because  acquiring a large number of labeled images is costly and time-consuming. The unavailability of a reasonable amount of data is one of the main obstacles that  prohibit the replication of similar advances for detecting other retinal vascular diseases. 
 
 The problem is  whether it is possible, and if so, how  to build upon
 the techniques and knowledge of DR detection for
 other retinal vascular diseases, given their limited amount of data?

\subsection{Our work}
In this paper, we propose and apply a transfer learning technique for retinal vascular disease detection.
The basic idea of transfer learning is to identify a well-studied source task that shares some similar features
with the target task for which there is limited data.   
Here we choose the well-studied DR detection as a source task, and transfer the learned knowledge to the early detection of retinopathy of prematurity (ROP), as the target task.
 
The transfer learning approach proposed here is different from the traditional 
 transfer learning approach widely adopted  for medical image analysis. 
 The former focuses on feature similarities between the source task and the target task, while the latter  uses a large natural and generic image data set such as ImageNet \cite{ILSVRC15} for pretraining, with the belief  that transfer learning from a large image data set helps improve the model performance.
Clearly, due to the large difference in image features,  effectiveness and robustness of this ImageNet-pretrained transfer learning vary and depend on the size of the pretraining data set and the size of the architecture \cite{raghu2019transfusion, mustafa2021supervised}.

To validate our DR-pretraining transfer learning approach, we  compare its performance  with the ImageNet-pretrained transfer learning approach, against the baseline results from the direct  training approach.  
To investigate the robustness of our approach, we conduct a series of experiments with  reduced training samples in the target task.

Our experimental results show the superior performance of the DR-pretrained approach,  not only in all metrics of AUROC, accuracy, precision, and sensitivity, but also in robustness. The robustness is with respect  to both the stochasticity in the training process and reduction in training samples. 

Our studies suggest the  effectiveness of our proposed transfer learning approach and its potential for a broad range of retinal vascular diseases or pathologies, where data is limited.

\subsection{Why Retinopathy of Prematurity?}

There are several reasons why 
ROP is chosen as the target task.

Firstly, ROP is a common retinal vascular disease. It is an abnormal blood vessel development in the retina of prematurely-born infants or infants with low birth weight \cite{Fiersone20183061}.
ROP can lead to permanent visual impairment and is one of the leading causes of infant blindness globally.  
It is  estimated that nineteen million children are visually impaired worldwide \cite{Blencowe2013}, among which ROP accounts for six to eighteen percent of childhood blindness \cite{gilbert_rahi_eckstein_osullivan_foster_1997}. Early treatment has  confirmed the efficacy of treatment for ROP \cite{10.1001/archopht.121.12.1684}. Therefore, it is crucial that at-risk infants receive timely retinal examinations for early detection of potential ROP. 

Secondly, early detection of ROP is particularly  challenging, due to  infants' inability of active participation in medical diagnosis. To minimize the number of
missed diagnoses for ROP  in infants, 
clinical screening  for ROP requires exceptionally high sensitivity scores. 

In light of these,  ROP presents itself as an ideal test bed for the feasibility of transfer learning technique utilizing feature similarities for
detecting retinal vascular diseases. And the success of transfer learning from DR to ROP, especially in comparison with existing approaches, is a barometer for the potential of this transfer learning technique.

\subsection{Related Work} 
As mentioned earlier, earlier works on medical image data analysis with transfer learning are mostly ImageNet pretrained, including 
 \cite{raghu2019transfusion}  in ophthalmology and radiology.
Their experiment shows that  ImagaNet-trained transfer learning offers little benefit to the performance of state-of-art models such as ResNet50 \cite{he2016deep} and Inception-v3 \cite{7780677}.
\cite{mustafa2021supervised} conducted similar and extended experiments at a larger scale in terms of the pretraining data set size and the architecture size.
The data set size ranges from 1.3 million (the standard ImageNet data set) to 300 million, and the number of parameters of the architecture ranges from 24 million (ResNet50) to 380 million.
Their results show that using larger architecture and a larger pretraining data set are the keys to benefit by transferring from natural image data sets.
 
 Very limited studies have been done beyond ImageNet-pretrained transfer learning, except for  \cite{chen2019med3d} which combines several data sets to develop a 3D lung cancer segmentation model, and \cite{Alzubaidi2021} which studies the transfer learning approach using an unlabeled pretraining data set.

To the best of our knowledge, our work is the first  that applies the supervised transfer learning method from one retinal vascular disease to another.

\section{Methodologies}
\label{sec:methods}
\subsection{Transfer Learning}
Transfer learning (TL) is a technique in machine learning which aims to transfer the learned knowledge from a domain to another
\cite{zhuang2020comprehensive}.
In transfer learning,  
\begin{itemize}
    \item a domain is a pair of a measurable space and a probability distribution on this space: $\mathcal{D} = (X, P(X))$, where
    $X$ is called a feature space and $P(X)$ is the distribution of the feature;
    \item a task in a domain $\mathcal{D}$ is a pair of a measurable space $Y$ and a function from $X$ to $Y$: $\mathcal{T}=(Y, f)$, where
    $Y$ is called the label space and $f$ is called a decision function.
\end{itemize}
The problem that machine learning tries to solve in a domain $\mathcal{D}$ for a task $\mathcal{T}$ is to learn $f$ from the samples drawn from $P(X)$.
Transfer learning utilizes the knowledge of a machine learning problem in a source domain $\mathcal{D}_s$ for a task $\mathcal{T}_s$ to improve performance of the learned decision function in a target domain $\mathcal{D}_t$ for a task $\mathcal{T}_t$.

Fine-tuning is the most popular approach when transfer learning is applied to deep learning.
In this approach, the network is first trained for the source task and then the weight is transferred to the target task.
Namely, given the source domain $\mathcal{D}_s=(X_s, P_s(X))$, source task $\mathcal{T}_s=(Y_s, f_s)$, and the network $\hat{f}(\cdot;\theta)$ with the network weight parameter $\theta$, transfer learning training is a bi-level optimization problem. The first step is to solve the following optimization problem
\begin{align*}
    \min_{\theta} \mathbb{E}_{x\sim P_s(X)}[loss(f_s(x), \hat{f}(x;\theta))].
\end{align*}
Suppose that an optimal weight $\theta_s^*$ is obtained from solving the above optimization problem, then the second step is to solve the following optimization problem
\begin{align*}
    \min_{\theta} \mathbb{E}_{x\sim P_t(X)}[loss(f_t(x), \hat{f}(x;\theta))],\ \theta_0=\theta_s^*.
\end{align*}
In other words, the trained network weight in the source task is used as the initial point of the optimization.
This approach is very popular in computer vision problems because the image features such as edges or corners are universal across the image domains, and the fine-tuning approach promotes the reuse of these learned features.

\begin{figure}[h]
    \begin{minipage}[b]{1.0\linewidth}
      \centering
      \centerline{\includegraphics[width=8.5cm]{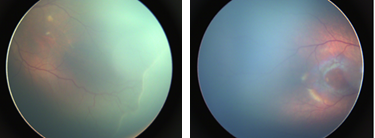}}
    \end{minipage}
    \caption{ROP positive sample (left) and negative sample (right); the white line on the right in the positive image is a disease feature called thickened ridge}
    \label{fig:rop_samples}
\end{figure}

\paragraph*{Target task.}
The target is to develop a deep neural network that correctly classifies input color fundus photograph (CFP) as ROP positive or ROP negative (Fig \ref{fig:rop_samples}).
In the transfer learning framework, the feature space is a space of 3D tensors, the feature distribution is the distribution of CFPs taken from infants, and the label space consists of ROP positive and ROP negative.

\paragraph*{Source tasks.}
There are two different source tasks:  one for the conventional ImageNet-pretrained  learning approach, and another for our proposed DR-pretrained transfer learning approach. 

In the first one, 
the feature space is (again) a space of 3D tensors, the feature distribution is the color natural images, and the label space consists of 1000 classes.

\begin{figure}[h]
    \begin{minipage}[b]{1.0\linewidth}
      \centering
      \centerline{\includegraphics[width=8.5cm]{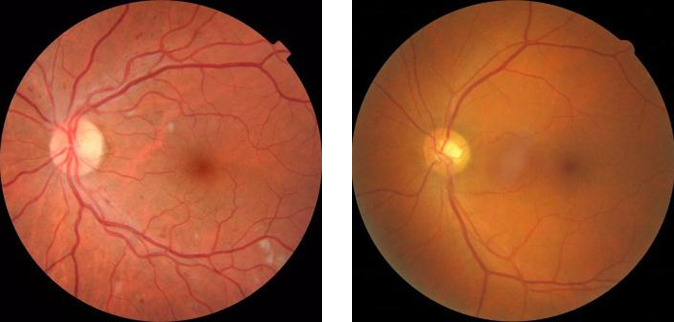}}
    \end{minipage}
    \caption{DR positive sample (left) and negative sample (right)}
    \label{fig:dr_samples}
\end{figure}

In the second one (Fig \ref{fig:dr_samples}),
  the feature space is (again) a space of 3D tensors, the feature distribution is the distribution of CFPs taken from diabetes patients, and the label space consists of DR positive or DR negative.

\section{Experiment}
We investigate the effect of transfer learning by comparing our DR-pretrained approach with the ImageNet-pretrained transfer learning, against the baseline results from the direct training with random initialization.
Throughout our experiment, ResNet50 architecture is used.

\paragraph*{Data collection.}
The CFPs taken from infants were collected from the Affiliated Eye Hospital of Nanchang University, which is an AAA (i.e., the highest ranked) hospital in China.
All images were de-identified according to patient privacy protection policy, and the ethics review was approved by the ethical committee of Affiliated Eye Hospital of Nanchang University (ID: YLP202103012).
The images were graded by 4 experienced ophthalmologists.
As a result, the data set consists of 9,727 images (2,310 positive samples, 7,417 negative samples).
The data set is randomly split into a training set and a test set by a ratio of 4:1.

The ImageNet data set consists of 1.3 million images collected online \cite{ILSVRC15}.
We used the public weight available in Keras \cite{chollet2015keras}.

The CFPs for DR were collected from the same hospital as the ROP data set, and the images were graded by experienced ophthalmologists whether DR positive or DR negative.
The DR data set consists of 36,126 images with 26,548 positive samples and 9,578 negative samples.

\paragraph*{Data augmentation.}
Each image in ROP data set or DR data set is applied brightness adjustment and random flipping.
Afterwards, each image is resized to 300$\times$300.

\paragraph*{Class rebalance.}
To mitigate the class imbalance issue in the DR and ROP data sets, we use a hybrid class balancing method:
First, the class weight in the loss function is set to be $1: r$ for the negative class to the positive class;
Second, when generating a minibatch, images from each class are sampled at the ratio of $r: 1$ so that each class has an equal impact on the training process.
The parameter $r$ is treated as one of the hyper parameters to be tuned.

\paragraph*{Metrics for performance evaluation.}
The trained models are evaluated by four metrics: the area under the receiver operator characteristic curve (AUROC), the accuracy, the precision, and the sensitivity.
To account for the stochastic nature of the training, each experiment is iterated three times with different random seeds and the metrics are averaged.

\paragraph*{Experiment with reduced training samples.}
To understand the effectiveness and robustness of transfer learning with limited data set, we further train the models with reduced training samples.
In this series of experiments, the same pretrained weights are used for each training i.e., pretraining data set is fully utilized, but the training samples in the target task is reduced by factors ranging from 0\% to 90\% with 10\% interval.
The test set is kept the same to ensure consistency for comparison.

\begin{figure}[h]
    \begin{minipage}[b]{1.0\linewidth}
      \centering
      \centerline{\includegraphics[width=6.5cm]{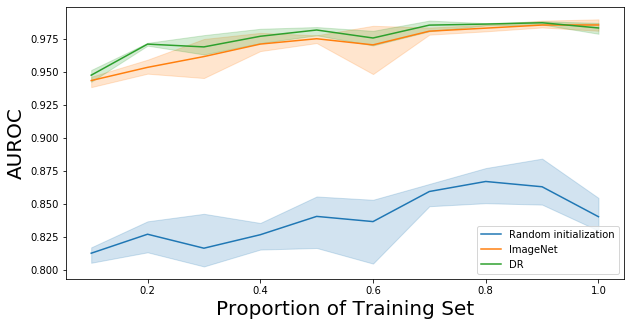}}
      \centerline{\includegraphics[width=6.5cm]{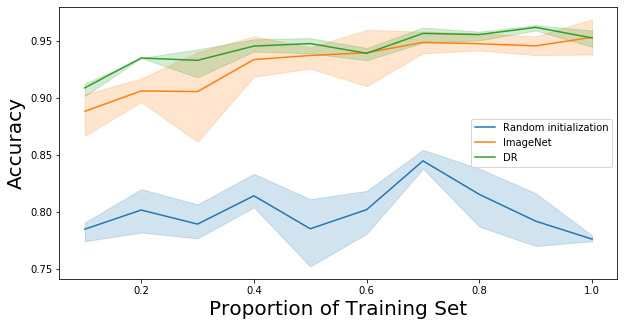}}
      \centerline{\includegraphics[width=6.5cm]{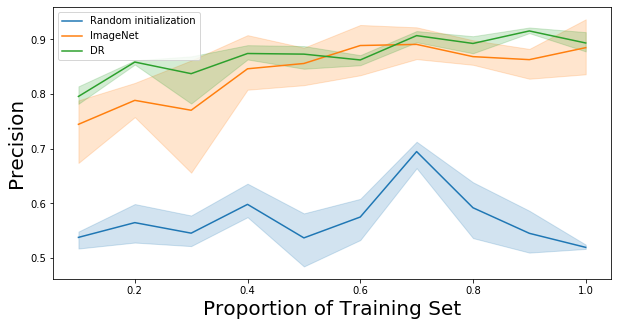}}
      \centerline{\includegraphics[width=6.5cm]{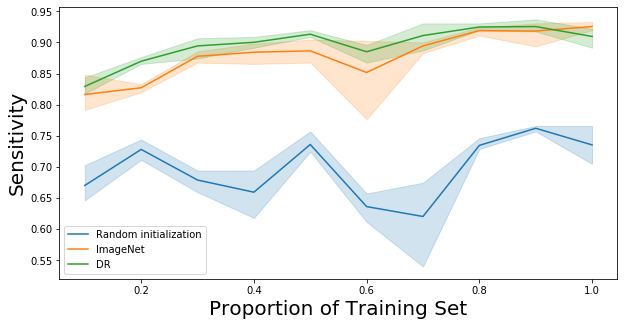}}
    \end{minipage}
    \caption{Changes in four metrics over training sample reduction, with the dark curves averaged over 3 experiments and the area around the curves showing the minimum and the maximum values in 3 experiments}
    \label{fig:res_plot}
\end{figure}

\paragraph*{Results.}
\label{sec:results}
The results are shown in Figure \ref{fig:res_plot} and tables \ref{table:meandiff},  \ref{table:std}, and \ref{table:reduction}.
We observe three critical advantages of our proposed approach via DR-pretraining over the traditional approach via ImageNet-pretraining.
Firstly, DR-pretraining demonstrates superior performance compared with ImageNet-pretraining.
Table \ref{table:meandiff} shows their mean percentage improvements from the direct training.
DR-pretraining dominates ImageNet-pretraining by all four metrics. 
Secondly, the DR-pretraining is more robust with respect to the stochasticity in the training process.
Table \ref{table:std} shows the mean percentage reduction of standard deviation from direct training.
DR-pretraining reduces the standard deviation by at least nearly 50\% for all metrics.
In contrast, ImageNet-pretraining adds more standard deviation (reduction of -46.6\%) in precision and shows almost no improvement (reduction of 2.94\%) in accuracy.
Lastly, DR-pretraining is more robust with respect to the reduction of training sample size.
The percentage changes of metrics from 100\% training size to 10\% training size are shown in Table \ref{table:reduction}.

These observations suggest 1) DR-pretraining dominates the traditional ImageNet-pretraining in all four metrics (AUROC, accuracy, precision, and sensitivity), 2) DR-pretraining is more robust with respect to both the stochasticity in the training process and   reduced training samples.

\begin{table}[h]
    \centering
    \begin{tabular}{|c|c|c|c|c|}\hline
        Pretraining & AUROC & Accuracy & Precision & Sensitivity \\ \hline
        DR & 16.4\% & 17.9\% & 53.5\% & 29.2\% \\ \hline
        ImageNet & 15.7\% & 16.3\% & 47.8\% & 26.9\% \\ \hline
    \end{tabular}
    \caption{Mean percentage improvement from direct training}
    \label{table:meandiff}
\end{table}

\begin{table}[h]
    \centering
    \begin{tabular}{|c|c|c|c|c|}\hline
        Pretraining & AUROC & Accuracy & Precision & Sensitivity \\ \hline
        DR & 75.4\% & 64.3\% & 47.5\% & 53.7\% \\ \hline
        ImageNet & 57.0\% & 2.94\% & -46.6\% & 25.1\% \\ \hline
    \end{tabular}
    \caption{Mean percentage reduction of standard deviation from direct training: the standard deviation calculated over different runs and the mean  calculated over different training sizes}
    \label{table:std}
\end{table}

\begin{table}[h]
    \centering
    \begin{tabular}{|c|c|c|c|c|}\hline
        Pretraining & AUROC & Accuracy & Precision & Sensitivity \\ \hline
        DR & 3.63\% & 4.61\% & 10.9\% & 8.82\% \\ \hline
        ImageNet & 4.26\% & 6.82\% & 15.8\% & 11.8\% \\ \hline
    \end{tabular}
    \caption{Percentage changes in all metrics from 100\% training size to 10\% training size}
    \label{table:reduction}
\end{table}

\section{Conclusion}
\label{sec:conclusion}
We propose a transfer learning  approach that uses the detection of well-studied retinal vascular disease as a source task to transfer the learned knowledge to the detection of an under-studied retinal vascular disease as a target task.
Experimental results demonstrate the superior performance of the DR-pretraining approach when compared with the traditional transfer learning and direct training approaches.
Our study shows promises of transfer learning  techniques utilizing feature similarities for general studies of retinal vascular diseases or other pathologies from different medical fields, where shortage of data is the main bottleneck for developing efficient deep-learning algorithms for medical image analysis.

\vfill\pagebreak

\bibliographystyle{IEEEbib}
\bibliography{main}

\end{document}